%% file: acl2019.tex
\documentclass[11pt,a4paper]{article}
\usepackage[hyperref]{acl2019}
\usepackage[utf8]{inputenc}
\usepackage{times}
\usepackage{helvet}
\usepackage{courier}
\usepackage{url}
\usepackage{hyperref}
\usepackage{times}
\usepackage{url}
\usepackage{latexsym}
\usepackage{wrapfig}
\usepackage{graphicx}
\usepackage{multirow}
\usepackage{breakcites}
\usepackage{latexsym}
\usepackage{booktabs}
\usepackage{tabularx}
\usepackage{tikz}
\usepackage{pifont}
\usepackage{amssymb}
\usepackage{amsmath}
\usepackage{subcaption}
\usepackage{enumitem}
\usepackage{floatrow}
\newfloatcommand{capbtabbox}{table}[][\FBwidth]
\usepackage{tablefootnote}

\newcommand{\ssymbol}[1]{^{\@fnsymbol{#1}}}

\aclfinalcopy % Uncomment this line for the final submission
\def\aclpaperid{9} %  Enter the acl Paper ID here

\title{Deep Retrieval-Based Dialogue Systems: A Short Review}

\author{Basma El Amel Boussaha, Nicolas Hernandez, Christine Jacquin, Emmanuel Morin \\
  LS2N, UMR CNRS 6004\\
  Université de Nantes, France \\
  \texttt{first.last@univ-nantes.fr} \\
  }

\date{}

\begin{document}
\maketitle
\begin{abstract}

Building dialogue systems that naturally converse with humans is being an attractive and an active research domain. Multiple systems are being designed everyday and several datasets are being available. For this reason, it is being hard to keep an up-to-date state-of-the-art. In this work, we present the latest and most relevant retrieval-based dialogue systems and the available datasets used to build and evaluate them. We discuss their limitations and provide insights and guidelines for future work.
%Today, retrieval-based dialogue systems are an active research area. In this survey paper, we present an overview of the latest works in this area, present the datasets and the metrics used in building and evaluating the existing systems. Currently, many works are interested in building and constructing dialogue systems and large datasets. In this work, we briefly overview the most recent systems and the available datasets and provide a catalogue for people 

\end{abstract}

\section{Introduction}

In the last two years, too many works from industry and academia were interested in building dialogue systems that can converse with humans in natural language by either generating responses \cite{vinyals2015neural, sordoni2015neural, serban2015building, D16-1127, wen2016network, zhang2018generating} or retrieving them from a set of candidate responses \cite{lowe-EtAl:2015:W15-46, wu2016response, wu2017sequential, Yang:2018:RRD:3209978.3210011, C18-1317}. Even if the generative systems can imitate humans and generate responses word by word, they suffer from the generality, non diversity and shortness of the generated responses \cite{li2015diversity, Gao:2018:NAC:3209978.3210183}. On the other hand, retrieval-based dialogue systems can produce coherent and syntactically correct responses, but they are constrained by the list of candidate responses. 

In this work, we are interested in studying retrieval-based dialogue systems as they have proved their efficiency in both academia and industry products such as the Alibaba's chatbot \textit{AliMe} \cite{P17-2079} and the Microsoft's social-bot \textit{XiaoIce}\footnote{\url{https://www.msxiaoice.com/}} \cite{Shum2018}. % The reasons why we write this paper are the following. (1) Because of the increasing interest in building dialogue systems, it is being hard to stay updated about the latest systems and the available datasets. (2) We noticed that different metrics and datasets are being used in the evaluation of these systems which makes their comparison not straightforward and we want to point this out. (3) We believe that it is important to briefly summarize the available information in one document and provide it to the researchers.
Our goal is to provide a short overview of the latest deep retrieval-based dialogue systems according to two aspects: novelty and relevance. Moreover, we describe the largest and most used public datasets and evaluation metrics. We also point out some drawbacks that we judge important to address in the future work. We believe that this work will be useful not only for researchers and developers recently interested in this research area but also researchers who are designing new systems so that they can easily find a recent and up-to-date list of the latest and most relevant systems. 

%The remainder of this paper is as follows. First, we describe the available approaches in Sections \ref{seq:appraoches} and \ref{seq:ensemble}. Then, we describe the available datasets in Section \ref{seq:datasets}. In Section \ref{seq:metrics}, we provide the evaluation metrics. Finally, in Section \ref{seq:conclusion}, we discuss the limitations of the systems and the datasets and provide insights and guidelines for future work.

\section{Retrieval-Based Systems} \label{seq:appraoches}

The existing retrieval-based dialogue systems belong to one of the following categories according to how they match the context with the response.

\subsection{Single-Turn Matching Models}

Systems of this category hypothesize that the response replies to the whole context. Thus, they consider the context utterances as one single utterance to which they  match the response without explicitly distinguishing the context utterances.

\noindent \textbf{Dual Encoder} The idea behind this model is to use a recurrent neural network to encode the whole context into a single vector \cite{lowe-EtAl:2015:W15-46}. First the context and the candidate response are presented using word embeddings and are fed into a LSTM \cite{hochreiter1997long} network. The last hidden state of the encoder is a vector that represents the context and the response. The response score is the similarity between the context and the response computed as the dot product between their two vectors and a matrix of learned parameters. Some variants of the dual encoder based on CNNs \cite{lecun1998gradient} and bidirectional LSTMs were also explored by \citet{kadlec2015improved}. Other similar single-turn matching models include \textbf{Attentive-LSTM} \cite{tan2015lstm}, \textbf{MV-LSTM} \cite{Wan:2016:MMR:3060832.3061030} and \textbf{Match-LSTM} \cite{N16-1170}. 

\textbf{ESIM} Enhanced Sequential Inference Model \cite{chen2019sequential} was originally developed by \citet{chen2018neural} for natural language inference. First, they concatenate the context utterances and following the same process as \citet{lowe-EtAl:2015:W15-46}, they encode the context and the response using a Bidirectional LSTM network. Then, cross attention mechanism is applied in order to model the semantic relation between the context and the response. Finally, max and mean pooling are applied and the output is transformed into a probability that the response is the next utterance of the given context using a multi-layer perceptron classifier.

\subsection{Multi-Turn Matching Models}

The main hypothesis of this category of dialogue systems, is that the response replies to each utterance of the context. Thus, the candidate response is matched with every utterance of the context. Then, an aggregation function is applied to combine the different matching scores and produce a response score. In the following, we present the most recent multi-turn matching systems.

\textbf{SMN} The Sequential Matching Network \cite{wu2017sequential} encodes separately the last 10 utterances of the context in addition to the response with a shared GRU \cite{chung2014empirical} encoder and obtain for each utterance and the response a matrix (all the hidden states of the encoder). This matrix represents the sequence information of each input. Moreover, a word similarity matrix is computed as a dot product between the matrices of each utterance and the response. These two matrices are used as input channels of a Convolutional Neural Network (CNN) followed by a max pooling that computes a two level matching vectors between the response and each context turn. A second GRU network aggregates the obtained vectors and produces a response ranking score.

\textbf{DAM} The Deep Attention Matching Network \cite{P18-1103} is an extension of the SMN \cite{wu2017sequential}. The DAM addresses the limitations of recurrent neural networks in capturing long-term and multi-grained semantic representations. This model is based entirely on the attention mechanism \cite{bahdanau2014neural}. It is inspired by the Transformer \cite{vaswani2017attention} to rank the response using self- and cross-attention. The first GRU encoder of the SMN model is replaced by five hierarchically stacked layers of self-attention. Five matrices of multi-grained representations of the context turns and the response are obtained instead of one matrix in the case of SMN. Following the same process of the SMN, the response matrices are matched with the context turns matrices and stacked together in a form of a 3D image (matrix). This image contains self- and cross-attention information of the inputs. Finally, a succession of convolution and max-pooling are applied on the image to produce the response score.

\textbf{DMN} The Deep Matching Network \cite{Yang:2018:RRD:3209978.3210011} extends the SMN\footnote{SMN is called DMN in their paper.} with external knowledge in two different ways. The first approach is based on the Pseudo-Relevance Feedback \cite{Cao:2008:SGE:1390334.1390377} named DMN-PRF and consists of extending the candidate response with relevant words extracted from the external knowledge (Question Answering (QA) data). The second approach incorporates external knowledge with QA correspondence Knowledge Distillation named DMN-KD. It adds a third input channel to the CNN of the SMN as a matrix of the \textit{Positive Pontwise Mutual Information (PPMI)} between words of the response and the most relevant responses retrieved from the external knowledge.

\textbf{DUA} The Deep Utterance Aggregation system \cite{C18-1317} also extends the SMN with an explicit weighting of the context utterances. The authors hypothesize that the last utterance of the context is the most relevant and thus concatenate its encoded representation with all the previous utterances in addition to the candidate response. After that, a gated self-matching attention \cite{P17-1018} is applied to remove redundant information from the obtained representation before feeding them into the CNN as in the SMN.

\begin{figure*}
\centering
\begin{subfigure}{.5\textwidth}
    \centering
  \def\svgwidth{0.7\textwidth}
\fontsize{8}{10}\selectfont
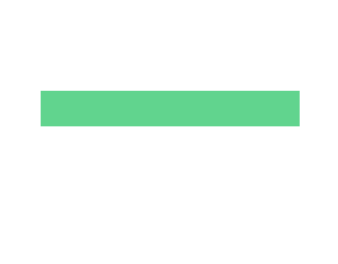
  \caption{Single-turn matching models}
  \label{fig:fig1}
\end{subfigure}%
\begin{subfigure}{.5\textwidth}
    \centering
  \def\svgwidth{0.7\textwidth}
\fontsize{8}{10}\selectfont
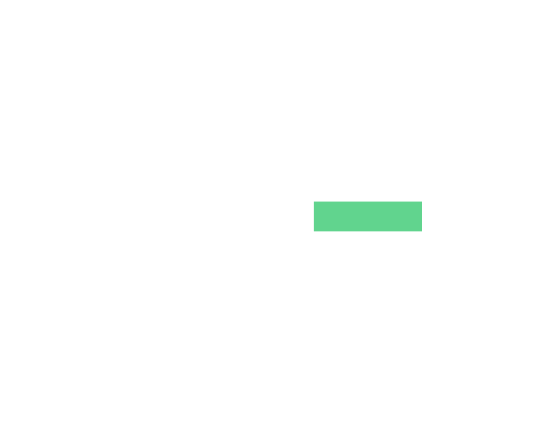
  \caption{Multi-turn matching models}
  \label{fig:fig2}
\end{subfigure}
\caption{General architectures of single- and multi-turn matching models}
\label{fig:architecture}
\end{figure*}

We summarize in Figure \ref{fig:architecture} the global architectures of single- and multi-turn systems.

\section{Ensemble\footnote{Sometimes called \textit{hybrid} systems.} Systems} \label{seq:ensemble}

In addition to the retrieval-based dialogue systems, generative dialogue systems were widely explored in the literature. Most of them are based on the \textit{sequence-to-sequence} (seq2seq) architecture \cite{sutskever2014sequence}. First, they encode the context into a vector, then decode this vector to generate the response word by word. Both categories have pros and cons. For instance, retrieval-based dialogue systems are constrained by a list of candidate responses and can only respond with one of the available responses. On the other hand, they can produce syntactically correct, diverse and long responses. Generative dialogue systems are not limited by a responses list and thus, they can generate more specific responses. However, they tend to generate "safe" responses which are short and general \cite{li2015diversity, serban2015building}. 

Recently, some studies were interested in combining both systems. \citet{P17-2079} built a hybrid system in which, for a given question (context), similar Question-Answer (QA) pairs are retrieved from a QA base using a retrieval system. Then, a ranker system computes a score for each retrieved answer A based on its related question~Q. Based on these scores, responses are ranked and the response with the highest score determines whether a new response is generated. If its score is higher than a threshold, this best response is returned. Otherwise, an \textit{attentive seq2seq} is used to generate a new response.

The system of \citet{ijcai2018-609} first retrieve candidate responses using the same previous process. Then, the query in addition to the retrieved responses are given as input to a generative system to produce a new response. Finally, the retrieved and the generated responses are ranked and the best response is returned. A retrieve and refine model was proposed by \citet{W18-5713}. First, it retrieves the best response and provides it, concatenated with the context, to an attentive seq2seq to generate a new response.

The most recent work of \citet{P18-1123} consists of an \textit{exemplar encoder-decoder} which first constructs a list of $k$-exemplar context-response pairs that are the most similar to the given context and response. Then, each exemplar context and response are encoded in addition to the original context. The exemplar responses vectors are concatenated with the original context vector and are fed into the decoder to generate a response. The score of the generated response is conditioned by the similarity between the exemplar contexts and the original context.

\section{Datasets} \label{seq:datasets}

Many publicly available datasets were used in evaluating most of the recent retrieval-based dialogue systems. In this section, we provide a non exhaustive list of the available datasets split into two categories (See appendix for further details).

\subsection{Negative Sampling Based Datasets}

This category regroups datasets where the negative candidate responses of each context were randomly selected without any human judgment. The most used dataset is the \textbf{Ubuntu Dialogue Corpus} (UDC) \cite{lowe-EtAl:2015:W15-46}. It is the largest available corpus, it contains Ubuntu related chat extracted from the freenode IRC. For each context, 10 candidate responses are provided among which, one response is the ground-truth and the rest are randomly sampled from the dataset. Today, three versions of UDC exist.

The third version \textbf{UDC (V3)} was released as part of the DSTC7 challenge with the \textbf{Advising} corpus \cite{DSTC7}. To the best of our knowledge, UDC (V3) and Advising are the only public datasets where each context has 100 candidate responses while almost all the other datasets provide 10 candidate responses. The Advising corpus contains teacher-student conversations collected at the University of Michigan with students playing teacher and student roles with simulated personas \cite{DSTC7}. The dataset includes additional information about preferences for workloads, class sizes, topic areas, etc.

The \textbf{MSDialog} \cite{Yang:2018:RRD:3209978.3210011} dataset was extracted from the Microsoft Answer Community\footnote{\url{https://answers.microsoft.com}} and consists of technical support conversations. As in UDC, negative sampling was used in order to produce the 9 negative responses for each context. Following the same negative response sampling process, the \textbf{E-commerce Dialogue Corpus} (EDC) was constructed \cite{C18-1317}. It is a public dataset which contains conversations between customers and customer service staff. 

\subsection{Human-Labeled Datasets}

Unlike the datasets of the first category, in order to construct the following datasets, humans were recruited to judge each candidate response and give it a label. Hence, every context may have more than one correct response.

The \textbf{Douban Conversation Corpus} \cite{wu2017sequential} contains human-human dialogues extracted from Douban\footnote{\url{https://www.douban.com/}} which is a popular social network in China. This is an open domain public dataset where conversations concern movies, books, music, etc. in contrast to the previously described datasets which are domain specific. 

\textbf{AliMe} data \cite{Yang:2018:RRD:3209978.3210011} is a human-machine corpus extracted from chat logs between customers and AliMe: the Alibaba chatbot \cite{P17-2079}. Analysts were asked to annotate candidate responses and assign positive labels to responses that match the context and negative labels otherwise. Unfortunately, the dataset is not public.

\section{Evaluation Metrics}\label{seq:metrics}

Evaluating dialogue systems is an open research problem \cite{lowe2017towards}. So far, information retrieval metrics have been widely used to evaluate retrieval-based dialogue systems. For instance Recall@k, Precision@k, Mean Recall Rank (MRR) and Mean Average Precision (MAP) evaluate the capacity of the dialogue systems to rank the correct response on top of the negative responses.

\section{Discussion and Conclusion} \label{seq:conclusion}

We presented the most recent retrieval-based dialogue systems and the available datasets. Even if these systems achieve good results, we believe that there are some drawbacks that should be addressed in the future. We summarize them in the following three points.

\noindent \textbf{Models} The existing models simply encode the context and the response and perform a semantic matching. No explicit modeling of dialogue acts, user intent and profile, etc. was performed. However, we believe that retrieving the next utterance of a dialogue depends on multiple parameters. Explicitly extracting and modeling these information while keeping the data-driven and end-to-end properties of the models could be of a great benefit. Moreover, we invite researchers to perform a qualitative error analysis. This can help understanding what is being captured by each architecture and what is being skipped to be addressed in the future.

The idea of ensemble systems is very interesting but until now, it has been done in one direction: the retrieval systems assist generative systems. We think that generating responses and then retrieving responses that match them can help the retrieval system in finding a better response. Hopefully this research area will be explored in the future. Furthermore, we would prefer to have the number of trainable parameters of each architecture to fairly compare two approaches in terms of their complexity. The source code of most of the state-of-the-art systems is available but they are implemented with different toolkits and use different data preprocessings (we refer to the Appendix). For this reason, we plan to build a single toolkit that implements the available approaches using the same libraries and provide a large panel of datasets and a unified choice of data preprocessing for an easy reproduction and a fair comparison.

\noindent \textbf{Datasets} As we can notice, among the six datasets that we presented in Section \ref{seq:datasets}, only one public dataset has been humanly labelled (Douban). We agree that human judgment is labour-intensive and subjective, but randomly sampling responses from the dataset and labeling them as negative responses is a very naive approach that may falsify the system training and evaluation. Moreover, most of the available datasets, limit the size of candidate responses to 10, except the latest datasets UDC (V3) and Advising released by DSTC7 \cite{DSTC7} which provide 100 candidate responses for each context. In practice, a retrieval-based dialogue system has to find the best response among a large set of candidate responses which is larger than 10. Unless another system is used to filter a smaller set of candidate responses to the response retrieval system, the size of candidate responses list should be larger.

\noindent \textbf{Evaluation Metrics} Recently, much effort has been done towards building more robust and smart systems but less effort has been done to define new evaluation metrics adapted to dialogue systems instead of IR and machine translation metrics \cite{serban2018survey}. We hope that researchers will orient some of their interests into this direction to provide more dialogue adapted metrics.

%In this work, we provided an overview of the most recent neural retrieval-based dialogue systems and the available datasets and evaluation metrics. We discussed the limitations that we judge important and gave insights for future works.

\newpage
\bibliographystyle{acl_natbib}
\bibliography{acl2019}

\end{document}

%% file: systems.pdf_tex
%% Accompanies image file 'systems.pdf' (pdf, eps, ps)
\begingroup%
  \makeatletter%
  \providecommand\color[2][]{%
    \errmessage{(Inkscape) Color is used for the text in Inkscape, but the package 'color.sty' is not loaded}%
    \renewcommand\color[2][]{}%
  }%
  \providecommand\transparent[1]{%
    \errmessage{(Inkscape) Transparency is used (non-zero) for the text in Inkscape, but the package 'transparent.sty' is not loaded}%
    \renewcommand\transparent[1]{}%
  }%
  \providecommand\rotatebox[2]{#2}%
  \newcommand*\fsize{\dimexpr\f@size pt\relax}%
  \newcommand*\lineheight[1]{\fontsize{\fsize}{#1\fsize}\selectfont}%
  \ifx\svgwidth\undefined%
    \setlength{\unitlength}{97.1242254bp}%
    \ifx\svgscale\undefined%
      \relax%
    \else%
      \setlength{\unitlength}{\unitlength * \real{\svgscale}}%
    \fi%
  \else%
    \setlength{\unitlength}{\svgwidth}%
  \fi%
  \global\let\svgwidth\undefined%
  \global\let\svgscale\undefined%
  \makeatother%
  \begin{picture}(1,0.78166295)%
    \lineheight{1}%
    \setlength\tabcolsep{0pt}%
    \put(1.02105901,-0.98764253){\color[rgb]{0,0,0}\makebox(0,0)[lt]{\begin{minipage}{0.9943083\unitlength}\centering \end{minipage}}}%
    \put(0.73865197,-0.81113808){\color[rgb]{0,0,0}\makebox(0,0)[lt]{\begin{minipage}{1.35908432\unitlength}\centering \end{minipage}}}%
    \put(1.26228179,-0.81702163){\color[rgb]{0,0,0}\makebox(0,0)[lt]{\begin{minipage}{1.02960924\unitlength}\centering \end{minipage}}}%
    \put(0,0){\includegraphics[width=\unitlength,page=1]{systems.pdf}}%
    \put(0.81372298,0.0754604){\color[rgb]{0,0,0}\makebox(0,0)[t]{\lineheight{0}\smash{\begin{tabular}[t]{c}Candidate\end{tabular}}}}%
    \put(0.18236185,0.07222288){\color[rgb]{0,0,0}\makebox(0,0)[t]{\lineheight{0}\smash{\begin{tabular}[t]{c}Context\end{tabular}}}}%
    \put(0.50415904,0.43836436){\color[rgb]{0,0,0}\makebox(0,0)[t]{\lineheight{0}\smash{\begin{tabular}[t]{c}Matching\end{tabular}}}}%
    \put(0.48645548,0.67190235){\color[rgb]{0,0,0}\makebox(0,0)[t]{\lineheight{0}\smash{\begin{tabular}[t]{c}Response\\\end{tabular}}}}%
    \put(0.8125327,0.02006903){\color[rgb]{0,0,0}\makebox(0,0)[t]{\lineheight{1.25}\smash{\begin{tabular}[t]{c}response\end{tabular}}}}%
    \put(0.48906503,0.61558957){\color[rgb]{0,0,0}\makebox(0,0)[t]{\lineheight{1.25}\smash{\begin{tabular}[t]{c}score\end{tabular}}}}%
    \put(0,0){\includegraphics[width=\unitlength,page=2]{systems.pdf}}%
    \put(0.17641068,0.25366586){\color[rgb]{0,0,0}\makebox(0,0)[t]{\lineheight{0}\smash{\begin{tabular}[t]{c}Encoder\end{tabular}}}}%
    \put(0,0){\includegraphics[width=\unitlength,page=3]{systems.pdf}}%
    \put(0.80962181,0.25366586){\color[rgb]{0,0,0}\makebox(0,0)[t]{\lineheight{0}\smash{\begin{tabular}[t]{c}Encoder\end{tabular}}}}%
    \put(0,0){\includegraphics[width=\unitlength,page=4]{systems.pdf}}%
  \end{picture}%
\endgroup%

%% file: systems_2.pdf_tex
%% Accompanies image file 'systems_2.pdf' (pdf, eps, ps)
\begingroup%
  \makeatletter%
  \providecommand\color[2][]{%
    \errmessage{(Inkscape) Color is used for the text in Inkscape, but the package 'color.sty' is not loaded}%
    \renewcommand\color[2][]{}%
  }%
  \providecommand\transparent[1]{%
    \errmessage{(Inkscape) Transparency is used (non-zero) for the text in Inkscape, but the package 'transparent.sty' is not loaded}%
    \renewcommand\transparent[1]{}%
  }%
  \providecommand\rotatebox[2]{#2}%
  \newcommand*\fsize{\dimexpr\f@size pt\relax}%
  \newcommand*\lineheight[1]{\fontsize{\fsize}{#1\fsize}\selectfont}%
  \ifx\svgwidth\undefined%
    \setlength{\unitlength}{160.12422524bp}%
    \ifx\svgscale\undefined%
      \relax%
    \else%
      \setlength{\unitlength}{\unitlength * \real{\svgscale}}%
    \fi%
  \else%
    \setlength{\unitlength}{\svgwidth}%
  \fi%
  \global\let\svgwidth\undefined%
  \global\let\svgscale\undefined%
  \makeatother%
  \begin{picture}(1,0.80362304)%
    \lineheight{1}%
    \setlength\tabcolsep{0pt}%
    \put(0.83478665,-0.35386851){\color[rgb]{0,0,0}\makebox(0,0)[lt]{\begin{minipage}{0.60310314\unitlength}\centering \end{minipage}}}%
    \put(0.66349111,-0.24680876){\color[rgb]{0,0,0}\makebox(0,0)[lt]{\begin{minipage}{0.82436003\unitlength}\centering \end{minipage}}}%
    \put(0.98110165,-0.25037746){\color[rgb]{0,0,0}\makebox(0,0)[lt]{\begin{minipage}{0.62451512\unitlength}\centering \end{minipage}}}%
    \put(0,0){\includegraphics[width=\unitlength,page=1]{systems_2.pdf}}%
    \put(0.88701261,0.11297611){\color[rgb]{0,0,0}\makebox(0,0)[t]{\lineheight{0}\smash{\begin{tabular}[t]{c}Candidate\end{tabular}}}}%
    \put(0.11998039,0.11101189){\color[rgb]{0,0,0}\makebox(0,0)[t]{\lineheight{0}\smash{\begin{tabular}[t]{c}$U_{1}$\end{tabular}}}}%
    \put(0.66143288,0.4022187){\color[rgb]{0,0,0}\makebox(0,0)[t]{\lineheight{0}\smash{\begin{tabular}[t]{c}Matching\end{tabular}}}}%
    \put(0.3419011,0.73704715){\color[rgb]{0,0,0}\makebox(0,0)[t]{\lineheight{0}\smash{\begin{tabular}[t]{c}Response\\\end{tabular}}}}%
    \put(0.88629022,0.07937798){\color[rgb]{0,0,0}\makebox(0,0)[t]{\lineheight{1.25}\smash{\begin{tabular}[t]{c}response\end{tabular}}}}%
    \put(0.34348387,0.70289032){\color[rgb]{0,0,0}\makebox(0,0)[t]{\lineheight{1.25}\smash{\begin{tabular}[t]{c}score\end{tabular}}}}%
    \put(0,0){\includegraphics[width=\unitlength,page=2]{systems_2.pdf}}%
    \put(0.1070029,0.25853794){\color[rgb]{0,0,0}\makebox(0,0)[t]{\lineheight{0}\smash{\begin{tabular}[t]{c}Encoder\end{tabular}}}}%
    \put(0,0){\includegraphics[width=\unitlength,page=3]{systems_2.pdf}}%
    \put(0.88452521,0.25853794){\color[rgb]{0,0,0}\makebox(0,0)[t]{\lineheight{0}\smash{\begin{tabular}[t]{c}Encoder\end{tabular}}}}%
    \put(0,0){\includegraphics[width=\unitlength,page=4]{systems_2.pdf}}%
    \put(0.35417346,0.11101189){\color[rgb]{0,0,0}\makebox(0,0)[t]{\lineheight{0}\smash{\begin{tabular}[t]{c}$U_{2}$\end{tabular}}}}%
    \put(0,0){\includegraphics[width=\unitlength,page=5]{systems_2.pdf}}%
    \put(0.34119603,0.25853794){\color[rgb]{0,0,0}\makebox(0,0)[t]{\lineheight{0}\smash{\begin{tabular}[t]{c}Encoder\end{tabular}}}}%
    \put(0,0){\includegraphics[width=\unitlength,page=6]{systems_2.pdf}}%
    \put(0.66330832,0.11101189){\color[rgb]{0,0,0}\makebox(0,0)[t]{\lineheight{0}\smash{\begin{tabular}[t]{c}$U_{n}$\end{tabular}}}}%
    \put(0,0){\includegraphics[width=\unitlength,page=7]{systems_2.pdf}}%
    \put(0.65033175,0.25853794){\color[rgb]{0,0,0}\makebox(0,0)[t]{\lineheight{0}\smash{\begin{tabular}[t]{c}Encoder\end{tabular}}}}%
    \put(0,0){\includegraphics[width=\unitlength,page=8]{systems_2.pdf}}%
    \put(0.35229832,0.43968968){\color[rgb]{0,0,0}\makebox(0,0)[t]{\lineheight{0}\smash{\begin{tabular}[t]{c}Matching\end{tabular}}}}%
    \put(0,0){\includegraphics[width=\unitlength,page=9]{systems_2.pdf}}%
    \put(0.10873717,0.47716066){\color[rgb]{0,0,0}\makebox(0,0)[t]{\lineheight{0}\smash{\begin{tabular}[t]{c}Matching\end{tabular}}}}%
    \put(0,0){\includegraphics[width=\unitlength,page=10]{systems_2.pdf}}%
    \put(0.37821211,0.59523881){\color[rgb]{0,0,0}\makebox(0,0)[t]{\lineheight{0}\smash{\begin{tabular}[t]{c}Aggregation\end{tabular}}}}%
    \put(0.42025571,0.00098861){\color[rgb]{0,0,0}\makebox(0,0)[t]{\lineheight{0}\smash{\begin{tabular}[t]{c}Context\end{tabular}}}}%
    \put(0,0){\includegraphics[width=\unitlength,page=11]{systems_2.pdf}}%
  \end{picture}%
\endgroup%